# Revenue Management without Demand Forecasting: A Data-Driven Approach for Bid Price Generation

Ezgi C. Eren[1] · Zhaoyang Zhang · Jonas Rauch · Ravi Kumar · Royce Kallesen

PROS Inc, Suite 600, 3200 Kirby Dr, Houston, TX 77098, USA

Traditional revenue management relies on long and stable historical data and predictable demand patterns. However, meeting those requirements is not always possible. Many industries face demand volatility on an ongoing basis, an example would be air cargo which has much shorter booking horizon with highly variable batch arrivals. Even for passenger airlines where revenue management (RM) is well-established, reacting to external shocks is a well-known challenge that requires user monitoring and manual intervention. Moreover, traditional RM comes with strict data requirements including historical bookings (or transactions) and pricing (or availability) even in the absence of any bookings, spanning multiple years. For companies that haven't established a practice in RM, that type of extensive data is usually not available.

We present a data-driven approach to RM which eliminates the need for demand forecasting and optimization techniques. We develop a methodology to generate bid prices using historical booking data only. Our approach is an ex-post greedy heuristic to estimate proxies for marginal opportunity costs as a function of remaining capacity and time-to-departure solely based on historical booking data. We utilize a neural network algorithm to project bid price estimations into the future.

We conduct an extensive simulation study where we measure our methodology's performance compared to that of an optimally generated bid price using dynamic programming (DP) and compare results in terms of both revenue and load factor. We also extend our simulations to measure performance of both data-driven and DP generated bid prices under the presence of demand misspecification. Our results show that our data-driven methodology stays near a theoretical optimum (<1% revenue gap) for a wide-range of settings, whereas DP deviates more significantly from the optimal as the magnitude of misspecification is increased. This highlights the robustness of our data-driven approach.

## 1. Introduction

Conventional revenue management (RM) systems rely on long and stable historical data to forecast demand, which depends on predictable demand patterns. However, susceptibility to external shocks is a well-documented challenge faced by RM practitioners (Weatherford, 2016). It is very hard for the forecast to quickly self-correct without user interference when faced with a sharp, sudden change in demand. These changes include shocks to demand intensity, willingness to pay of customers, and/or booking patterns. The recent COVID-19 pandemic is an extreme example of a demand shock, where many industries including airline (Garrow & Lurkin, 2021) and hospitality (Guillet & Chi Chu, 2021) faced an unprecedented decline in demand in addition to change in booking patterns. In contrast, other industries, such as air cargo, observed a surge in demand given the dramatic increase in online consumption (Ozden & Celik, 2021). Apart from rare global disasters, demand shocks and changes in demand patterns occur frequently at smaller scales and more local scopes. These sudden disruptions to demand volume and patterns curtail the effectiveness of the RM and pricing controls in place. This is one of the main reasons that a close monitoring and influencing of forecasting output is a common practice in RM (Weatherford, 2016).

[1] Corresponding author: eeren@pros.com

Some industries face demand volatility on an ongoing basis. Air cargo is a good example of this with its so-called "late and lumpy" demand (Mardan, 2010). The booking horizon is very short with most of the requests happening within the last few days before departures, hence the *late* arriving demand. Moreover, shipment sizes have a lot of variability in both volume and weight and are often *lumpy*. For such industries with extreme demand volatility, generating a demand forecast with an acceptable accuracy is often deemed infeasible.

Conventional RM also has extensive data requirements, and most of them are related to demand forecasting, which is the first step of RM where demand distribution parameters are estimated from historical transactions and forecasted. Demand forecasting requires historical booking data, usually long enough to capture repetitive (e.g., seasonal) patterns. As the historically observed demand is already a function of historical pricing, availability and capacity limits**,** forecasting frequently includes an unconstraining (a.k.a. uncensoring) methodology. This is required to estimate the unconditional demand or demand as a function of pricing implemented in future (Walczak, Boyd, & Cramer, 2011)**.** Hence, for the entire timespan of historical booking data used, there is a need to have the corresponding historical pricing and/or availability. This usually corresponds to a vast amount of data, and is often not readily available if the industry or company is relatively new to RM.

Finally, in the second step of conventional RM, an optimization methodology is used to compute the optimal pricing or availability (which we will refer to as "controls") given the forecasted demand and future capacity. Therefore, an input or estimation of future capacity (e.g., schedule data for airlines) is another required data component. The *optimal* controls generated from this optimization step are then used to accept/reject future incoming demand or to set pricing decisions. This in turn constrains the future bookings, creating a continuous feedback loop. Hence, it is quite possible that any model misspecification in demand forecasting could result in a vicious cycle of deteriorating forecast and revenue loss, such as the spiral down effect seen in the case of systemic low bias (Walczak, Boyd, & Cramer, 2011). Given the sensitivity of the RM system's performance to its input data, there is a vast amount of literature on RM related to efforts to improve forecasting (see Weatherford (2016) for a comprehensive overview).

In this paper, we pose a completely different question of interest and a new approach to solve the RM problem: Is it possible to directly prescribe control parameters of the RM system using historical data *without any demand forecasting?* We propose a novel methodology that does not rely on the abovementioned framework with demand forecasting and optimization. It utilizes historical booking data to directly generate the required RM output, skipping demand forecasting altogether. Our approach also alleviates the need of extensive historical data and eases the implementation of RM in practice. That makes our proposed framework accessible to industries relatively new to RM in addition to industries with demand volatility. We focus on generation of marginal opportunity costs (*bid prices),* which are an important and commonly used output of an RM system, as we will review later in this section. Through a comprehensive numerical study, we show that our method generates a robust bid price output even in the scenarios of unreliable historical data. Some unique characteristics of our approach that set it apart from conventional RM methodologies, can be summarized below:

- *Direct:* We generate bid prices directly from historical booking data, skipping demand forecasting completely.
- *Adaptive:* As we illustrate with an extensive simulation study in Section 4, our methodology is robust with good revenue performance under various scenarios with mis-specified demand.
- *Machine Learning (ML) Based:* We utilize a deep Neural Network (NN) model as the predictive algorithm, which provides flexibility with respect to data sources and features used for prediction. It is well recognized that incorporating additional factors such as competitor data, shopping data, and market indicators can improve prediction accuracy in RM and pricing (Gautam, Nayak, & Shebalov, 2021). Although our approach does not require any inputs other than the historical booking data, it can easily incorporate additional data when available.

## 1.1 Bid Prices in RM

There are many reasons that the use of bid prices in RM systems is so popular (Talluri & Van Ryzin, 2004). Bid prices are revenue-based and have an intuitive economical interpretation as they represent the marginal opportunity cost associated with losing capacity. For a bid price to be effective, it should capture both the impacts of capacity (e.g., number of seats remaining on a flight leg) and time (e.g., time-to-departure) dimensions. This requires storing at least a vector of bid prices indexed by remaining capacity and valid for the current point in time. Bid price controls are therefore easy to implement, because all that is required in real-time is a look-up: For an incoming booking request at any given time, one simply needs to extract the bid price values corresponding to current remaining capacities of all the requested resources and sum over them to get the total bid price associated with the incoming request. The total bid price is then used as input for an availability control or dynamic pricing engine. Finally, bid price controls have shown to have very good revenue performance and they are theoretically near optimal asymptotically as both demand and capacities scale up. Computing the right bid price and implementing it correctly is critical to realizing their revenue benefits.

The term *bid price* has multiple conflicting definitions even when confined to the RM literatures, hence we will first set the stage by defining what we mean by a bid price. In this paper, we refer to the term *bid price* primarily as an *opportunity cost* in the context of RM. It represents *the expected loss in future revenue from using the capacity now rather than reserving it for future use.* Equivalently, bid price can be defined as the marginal value of capacity. This definition is also in line with *expected marginal seat revenue (EMSR)* (Belobaba, 1987; Belobaba, 1989) which is indeed the first attempt to calculate a simplified bid price in the context of RM under certain assumptions. Readers can refer to McGill & Van Ryzin (1999) for a list of common RM terminology, including bid price, EMSR and opportunity cost. In the context of bid-price based controls, bid prices serve as more of a threshold price (which may imply some adjustment to the opportunity cost or the fares that are being compared to them), and do not need to be equal to the opportunity cost to generate near-optimal revenue. However, capturing the opportunity cost correctly is still desirable as bid price output from an RM system can be utilized in other decision making downstream, including but not limited to pricing (Talluri & Van Ryzin, 1998; Talluri & Van Ryzin, 2004). As the disentanglement of capacity control and pricing becomes more favorable with the increased feasibility of continuous/classless pricing (Rauch, Isler, & Pölt, 2018; Liotta, 2019), capturing the opportunity cost from the RM system becomes an even more critical input into the pricing module. Although we primarily focus on generating a bid price that would work in a continuous pricing setting, this assumption is not limiting and does not prevent the generated bid price to be valid in a class-based setting. To summarize, regardless of the setting, bid prices from an RM system are an essential component of availability and pricing decisions under capacity constraints.

The paper is organized as follows: In Section 2, we review the related literature and show how our approach fits in. We introduce the terminology, assumptions, and the detailed methodology in Section 3, after a brief review of the relationship to existing bid price and EMSR generation techniques. In Section 4, we present an extensive numerical study to test our proposed methodology with respect to an optimal benchmark. In this section, we also present a detailed empirical analysis of the results from simulations where the underlying demand intensity is mis-specified. This tests the robustness of our methodology. Finally, we conclude with a summary of contributions and a discussion of potential future work in Section 5.

## 2. Literature Review

Since there is no demand forecasting in our proposed approach, the required assumptions on demand distributions which are essential to conventional RM become irrelevant. Hence, we will review the recent trending literature in data-driven RM and robust RM, as our approach is related to those two fields, and can be placed under the data-driven RM literature. Both fields represent RM literature that explores the possibility of generating RM controls without relying on a demand forecast, even though they diverge in their approaches and the conditions for viable applications.

Van Ryzin & McGill (2000) develop an adaptive algorithm to generate and update airline seat protection levels. Their method utilizes historical data of seat-filling event fequencies for each booking class to create a feed-back loop to update the protection levels for a single-leg setting. Their approach can be regarded as the pioneering work for data-driven RM (a.k.a data-driven optimization, distribution-free RM). These methods generate and update control output without assuming a probability distribution or arrival process for the incoming demand, but they still rely on historical data, albeit with less strict requirements.

Following Van Ryzin & McGill (2000), there have been a number of studies that can be categorized under data-driven RM. Ball & Queyranne (2009) is another seminal paper that focuses again on booking class limits for the single-leg problem with no demand forecast and assumptions on demand distribution. Their methodology is motivated by high-uncertainty settings. They derive a competitive ratio with respect to a clairvoyant optimum for various policies including bid price controls. And they assume independent demand between fare classes. Ma (2010) extends Ball & Queyranne's results to a priceable demand setting where the arrivals are characterized by their willingness-to-pay, or so-called valuation, and the control policy is reduced to dynamic pricing. Ma derives inventory-dependent price skimming policies along with extended competitive ratio results. Ma's proposed methodology is called valuation tracking, which is a heuristic approach based on tracking the possible values for the optimum and following the most inventory-conservative control that maintains the desired competitive ratio.

Chen & Farias (2013) propose data-driven dynamic pricing where the demand forecast is unspecified or partially specified with the scale of demand intensity being modulated by an exogenous market size stochastic process. They extend Gallego & Van Ryzin (1994)'s fixed price policy to a re-optimized fixed price policy (RFP-$\Delta$) to handle demand uncertainty by frequent re-optimization and re-estimation of the instantaneous market size. Their approach relies only on easily accessible data such as sales data. Chen & Farias compare their policy's results to a clairvoyant optimal policy that has a perfect knowledge of the realization of customer demand over time, and they prove that the relative loss is bounded under certain assumptions.

Bertsimas & Vayanos (2015) studies data-driven dynamic pricing in the context of demand learning, requiring again only historical sales data. Despite their assumptions being the most applicable to a retail setting, the extension to network RM problem is discussed in the scope of the paper. Their methodology focuses on maximizing the performance of the worst-case scenario in a data-driven uncertainty region, so their work can be categorized in the intersection of data-driven RM and robust optimization, which we will review next. The uncertainty region is updated on-the-fly during optimization, which essentially unifies the estimation and optimization processes. Their pricing strategy strikes an optimal balance between exploitation and exploration to reduce uncertainty while obtaining good revenue performance.

Data-driven methodologies still rely on historical data even with little-to-no assumptions on the underlying demand process. Contrary to data-driven RM, the related domain of robust optimization does not require historical data in addition to being light on demand assumptions. Under this line of research, Perakis & Roels (2009) study the network RM problem with unknown demand and no historical data. They define an uncertainty region to represent minimum and maximum possible demand. And they develop linear program (LP) and mixed integer program (MIP) formulations to derive optimal policies under the objectives of minimizing maximum regret or maximizing minimum revenue or profit. These approaches are well suited to optimize the worst case scenarios under unstable business environments.

There are also studies that focus on network RM problem with a data-driven approach. Akan & Ata (2009) and Ata & Akan (2015) build and analyze a perturbed version of the network LP with no assumptions on the demand distribution. They prove that the optimal policies are of the bid-price type, and establish the conditions for near optimality. In addition to optimality, their characterization of bid prices as martingale processes is a novel approach that leads to useful practical insights. Although their model still needs a demand-estimate input, their setting is light on assumptions as demand is modeled as random diffusions. Jasin (2015) also considers the standard network RM problem and studies the performance of an LP-based control, the probabilistic allocation control (PAC), in the presence of unknown demand parameters. Jasin also shows the importance of the frequency of re-optimization and re-estimation on revenue performance.

Finally, it is worth mentioning the related line of data-driven research on the classical newsvendor problem. Ban & Rudin (2018) study a newsvendor problem with unknown demand distribution given historical data possibly with features included. They extend the Sample Average Approximation (SAA) approach from Retsef et al. (2007) to include features and derive LP formulations both for small data and big data (i.e., data with large number of features). Readers can refer to Phillips (2015) for a list of papers on the data-driven newspaper as well as data-driven RM.

Our proposed methodology falls under data-driven RM techniques, as it does not make any assumptions on demand in addition to skipping demand estimation completely. We differentiate our work from the already existing data-driven RM approaches, by generating a dynamic bid price rather than a fixed bid price, as a function of remaining capacity as well as the remaining time (e.g., time-to-departure). Our work is not as closely related to the robust optimization literature, as it relies on historical data and does not leverage mathematical optimization techniques as a solution method. However, it is worth mentioning that our numerical results show strong robustness to the instabilities in the incoming data. Finally, even though we primarily focus on the single-leg (or single-resource) setting in this paper, our methodology is still applicable to the network RM problem as will be discussed in Section 3.

# 3. Problem Description and Methodology

Before we describe the problem and methodology in detail, we would like to review EMSR and dynamic-programming (DP) based bid prices. This will facilitate better description of the motivation and connection to these existing RM methodologies and their generated outputs. We will use airline terminology throughout the problem description, as the origination of these methods from airline RM makes it a suitable candidate to describe the problem without limiting its broader applicability.

## 3.1 Expected Marginal Seat Revenue (EMSR)

There are few ways to estimate expected marginal revenue, and we refer the reader to (Belobaba, 1987; Belobaba, 1989) for a complete presentation of methodology for the two commonly used heuristics, EMSR-a and EMSR-b, which build on top of Littlewood's rule as we will review first.

Let's assume two product classes, with prices $f_1 > f_2$, with corresponding demand variables $D_1, D_2$, and $s$ seats remaining on a given flight. Littlewood's rule states that one should only sell the lower fare class when $f_2 \geq f_1 P(D_1 > s)$, in other words, if its fare is greater than or equal to the expected marginal revenue of $s^{th}$ seat. This threshold is essentially a bid price as defined in Section 1 and is equal to the higher fare class's fare multiplied by the probability of getting at least s passengers for that class

Note that this estimation requires a probability distribution of demand, which implies the need for forecasting including an unconstraining method as mentioned in Section 1. That unconstraining method would require historical controls that specifies when the two classes were open, in addition to historical bookings.

Belobaba (1987, 1999) extends the Littlewood's rule to handle multiple fare classes with the popular EMSR methods mentioned above. These methods also use fixed fares associated with fare classes and assume a probability distribution for each fare class's demand, $D_i$. Typically, the probability distribution for each class's demand is estimated based on historical booking information, which again implies the need for forecasting with an unconstraining method.

Our method takes an empirical data-driven approach by only using actual transactions without assuming or forecasting probability distributions of underlying demand. Also, the dependance of EMSR approaches on the fare class structure is limiting, as it is not applicable to industries other than the airline that do not operate with fare classes or fare buckets. Next, we will review bid price generation via DP under a continuous pricing setting without fare classes.

## 3.2 DP Generated Bid Price

We follow similar setting and formulation as in Gallego & van Ryzin (1994). The demand for each departure date is modeled as a non-stationary Poisson arrival process with a mean arrival rate $\lambda(t)$, where $t$ represents the time-to-departure. We consider priceable demand (Boyd & Kallesen, 2004), where the probability of purchase at a given price $p$ at time $t$ is denoted by $P_w(p,t)$. $P_w(p,t)$ is the probability of WTP of the arriving customer at time $t$ being greater than offered price, $p$. The booking horizon is divided into intervals of length $\delta t$ such that the probability of two arrivals happening in each interval is negligible. Defining the state at any given time $t$ as the remaining inventory $x$ (i.e., number of seats in the leg), we can model the pricing process via a DP formulation as follows:

$$V(x,t) = \max_{p}\{\lambda(t)\delta t \cdot P_w(p,t)\,[p + V(x-1,t-\delta t)] + [1-\lambda(t)\delta t \cdot P_w(p,t)]V(x,t-\delta t)\}. \quad (1)$$

Here $V(x,t)$ is the value function that represents the expected revenue-to-come from future sales given $x$ units of remaining inventory and $t$ time units to departure. After rearranging terms, Equation (1) can be rewritten as

$$V(x,t) = \max_{p}\{\lambda(t)\delta t \cdot P_w(p,t)\,(pb(x,t-\delta t))\} + V(x,t-\delta t), \quad (2)$$

where

$$b^*(x,t-\delta t) = V(x,t-\delta t) - V(x-1,t-\delta t) \quad (3)$$

The term $b^*(x,t)$ represents the optimal bid price with $x$ units of remaining inventory ($x$ seats) and t time units remaining to departure. In line with our definition of opportunity cost, Equation (3) provides a mathematical representation of how bid price measures exactly the expected revenue loss from having one less seat at a given inventory level and time-to-departure, and can be solved iteratively using the boundary conditions given by $V(x,0) = 0\ \forall x$, and $\mathrm{V}(0,t) = 0\ \forall t$. Conceptually, $b^*(x,t)$ is related to EMSR of the $x^{th}$ unit of resource (seat), with EMSR not capturing the time component.

Following the DP formulation above, it is obvious that optimal bid price, $b^*(x,t)$, is an important determinant of optimal price, $p^*(x,t)$. We will not go into the details of that relationship here, but a derivation of optimal price as a function of bid price under exponential demand assumption is provided in Appendix A. In summary, an estimation of demand volume parameter and WTP distribution (which is a function of purchase probability) are required to generate both $b^*(x,t)$ and $p^*(x,t)$. This is typically done with a forecasting algorithm using the historical demand data and the assumptions listed above. Equation (3) provides the theoretical optimal when those parameters are estimated with 100% accuracy, which is never feasible in practice. Therefore, parameter and model misspecifications frequently result in suboptimal decisions (Nambiar, Simchi-Levi, & Wang, 2019).

The bid prices generated by Equation (3) have nice structural properties that we leverage as the underlying logic behind our proposed approach. For any fixed remaining inventory level, $x$, the optimal bid price $b^*(x,t)$ increases as time-to-departure, $t$, increases. Similarly, for any fixed time remaining to departure, $b^*(x,t)$ decreases as remaining inventory level, $x$, increases. These monotonicity properties are intuitive, as the inventory tends to become more valuable as the remaining inventory level decreases, and it becomes less valuable as the departure time approaches. As remaining inventory usually keeps decreasing as the departure date approaches, these two opposing effects are expected to counterbalance each other to produce stable bid price values along the booking horizon. Talluri & Van Ryzin (1998) prove this property for asymptotically optimal bid prices in the network setting, and Ata & Akan (2015) establish the martingale property of optimal bid prices in a more generalized network setting.

## 3.3 Data-Driven Approach for Bid Price Generation

Our proposed approach utilizes historical booking data to directly estimate the value of each unit of inventory (e.g., seat) at any given time-to-departure. Before going into the details of our methodology, we begin with problem description.

### 3.3.1 Problem Description

Consider a single flight leg across multiple departures over time. Let $\boldsymbol{P} = \{P^1, P^2, \dots, P^{|I|}\}$ and $\boldsymbol{T} = \{T^1, T^2, \dots, T^{|I|}\}$ represent the sequence of vectors corresponding to the booked prices and the time-to-departure for each flight $i \in I = \{1,2, \dots, |I|\}$ in the historical booking data. The components of the vectors $P^i = [p_1^i, \dots, p_n^i, \dots, p_{N^i}^i]^\top$ and the vectors $T^i = [t_1^i, \dots, t_n^i, \dots, t_{N^i}^i]^\top$ hold the price and time information for all bookings: For the $i^{th}$ flight, $N^i$ represents the total number of bookings, and $p_n^i$ is the price, and $t_n^i$ is the time-to-departure for the $n^{th}$ booking on that flight. Note that $t_1^i > t_2^i > \cdots > t_{N^i}^i$ by definition. It is common to record the time-to-departure in an aggregated form (e.g., grouping the time-to-departure into days-priors to the departure date), so the data in $\boldsymbol{T}$ can be in any time units or clustering of time periods.

Our objective is to generate a bid price matrix, $\boldsymbol{B} = [b(x,t)]$, from historical data $\boldsymbol{P}$ and $\boldsymbol{T}$, where $b(x,t)$ is the bid price, or the opportunity cost for remaining capacity of $x$ at the remaining time of $t$.

### 3.3.2 Methodology

We first introduce the *observation building* step which is core to our proposed methodology. The objective of this step is to transform the historical booking data into a proxy of bid prices, which is then used to train a neural network to fix its parameters so the model can be used to estimate future bid prices.

#### *3.3.2.1 Observation Building*

To illustrate the fundamental idea behind our approach, we first begin with considering only the prices from historical data, $\boldsymbol{P}$, ignoring time component, $\boldsymbol{T}$, for now. Consider a single flight $i$ from history with capacity $C$, and the corresponding sales data for the flight, $P^i = [p_1^i, \dots, p_n^i, \dots, p_{N^i}^i]^\top$. If we were to estimate the revenue opportunity considering only one seat in inventory, which price would we have accepted for that seat given the full information post-flight? As the goal is to maximize revenue, it would be the highest price/fare from $P^i$. Similarly, if the remaining capacity was only two seats, the two transactions with the highest prices/fares from $P^i$ would be picked. If the total capacity (i.e., number of seats) exceeds total number of bookings, $C > N_i$, the revenue opportunity for any number of remaining seats in the range $\{N_i + 1, \dots, C\}$ is set to zero, as seats which are not booked simply generated zero value. Following this logic, one can transform $P^i$ into a new sorted vector, $\tilde{P}^i$, where the prices are simply ordered in decreasing order, and any remaining capacity up to $C$ is filled with zero. This process would be applied to all flights in $\boldsymbol{I}$ independently. A numerical example is provided below.

Example 1: Assume $C = 5$ and $\boldsymbol{I} = \{1,2\}$. For the given pricing data $P^1 = [80, 70, 90]^\top$ and $P^2 = [80\ 90\ 70\ 70]^\top$, the transformed observation vectors are $\tilde{P}^1 = [90, 80, 70, 0, 0]^\top$ and $\tilde{P}^2 = [90, 80, 70, 70, 0]^\top$. To train a model to predict these values, we also need the corresponding covariates for each data point, which, in this case, consists of only the remaining capacity index. The training data for an ML model in this case would look like this:

$$X = \begin{bmatrix} 1 \\ 2 \\ 3 \\ 4 \\ 5 \\ 1 \\ 2 \\ 3 \\ 4 \\ 5 \end{bmatrix}, Y = \begin{bmatrix} 90 \\ 80 \\ 70 \\ 0 \\ 0 \\ 90 \\ 80 \\ 70 \\ 70 \\ 0 \end{bmatrix}.$$

We provide a comparison of EMSR and our methodology's results using a simple average as predictive algorithm after observation building with a numerical example in Appendix B: Numerical Example for Relation of Data-Driven Approach to EMSR, for better understanding of the relation between the two approaches and proximity of results.

Even though we started with the connection to EMSR, note that the above transformation is in line with the monotonicity property of bid prices explained in Section 3.2; and the total revenue generated from a flight $i$, $\sum_n p_n^i$, can be considered as an estimator for the value function defined in Equation (1).

Next, we will extend the methodology to include the time-to-departure dimension, similar to the bid price function we reviewed in Section 3.2. For this, we begin by grouping booking horizon into time periods depending on similarity of booking volumes. It is common practice to set up data collection points (DCPs) which divide booking horizon into discrete intervals that strike for a right level of aggregation (Walczak, Boyd, & Cramer, 2011). Let us define the DCPs for a given historical data as $D = [d_1, d_2, \ldots, d_{|D|}]$ in time units to departure, with $d_1 > d_2 > \cdots > d_{|D|}$. For each DCP, $d_j \in D$, we construct a subset of prices $P_{\leq d_j}^i$ from the vector $P^i$, such that $P_{\leq d_j}^i = \{P_n^i \mid T_n^i \leq d_j\}$. Hence, we consider the subset of prices that were realized from DCP $d_j$ until departure. This is in line with the fact that to determine the bid price (or revenue opportunity of any remaining capacity) at a given DCP $d_j$, all future demand from that point on needs to be considered following the DP formulation provided in Section 3.1. After constructing $P_{\leq d_j}^i$ for all $d_j \in D$, we transform each $P_{\leq d_j}^i$ into a vector $\tilde{P}_{\leq d_j}^i$ by sorting and zero-padding as described above.

Example 2: Assume $C = 5$ and $\boldsymbol{I} = \{1,2\}$, $\boldsymbol{P} = \{[90\ 70\ 80], [80\ 90\ 70\ 70]\}$ as in Example 1. Now, we also include the time-to-departure data for $\boldsymbol{P}$, given in days by $\boldsymbol{T} = \{[2\ 1\ 0], [2\ 2\ 1\ 0]\}$. For the sake of simplicity, we assume each days-prior to departure constitutes a separate DCP, hence $D = [2,1,0]$. The transformed matrices for each flight are given by,

$$\tilde{P}^1 = [\tilde{P}_{\leq 2}^1 \quad \tilde{P}_{\leq 1}^1 \quad \tilde{P}_{\leq 0}^1] = \begin{bmatrix} 90 & 80 & 80 \\ 80 & 70 & 0 \\ 70 & 0 & 0 \\ 0 & 0 & 0 \\ 0 & 0 & 0 \end{bmatrix}, \tilde{P}^2 = [\tilde{P}_{\leq 2}^2 \quad \tilde{P}_{\leq 1}^2 \quad \tilde{P}_{\leq 0}^2] = \begin{bmatrix} 90 & 70 & 70 \\ 80 & 70 & 0 \\ 70 & 0 & 0 \\ 70 & 0 & 0 \\ 0 & 0 & 0 \end{bmatrix}.$$

Note that for each DCP, the prices that were incurred on and after that DCP are used. They are sorted in decreasing order and indexed accordingly. Again, in order to train an ML model, we arrange all values in a single column vector and create a corresponding covariate matrix $X$ that contains for each data point the remaining capacity and DCP index:

$$X = \begin{bmatrix} 2 & 1 \\ 2 & 2 \\ 2 & 3 \\ 2 & 4 \\ 2 & 5 \\ 1 & 1 \\ 1 & 2 \\ 1 & 3 \\ 1 & 4 \\ 1 & 5 \\ 0 & 1 \\ 0 & 2 \\ 0 & 3 \\ 0 & 4 \\ 0 & 5 \\ 2 & 1 \\ 2 & 2 \\ 2 & 3 \\ 2 & 4 \\ 2 & 5 \\ 1 & 1 \\ 1 & 2 \\ 1 & 3 \\ 1 & 4 \\ 1 & 5 \\ 0 & 1 \\ 0 & 2 \\ 0 & 3 \\ 0 & 4 \\ 0 & 5 \end{bmatrix}, Y = \begin{bmatrix} 90 \\ 80 \\ 70 \\ 0 \\ 0 \\ 80 \\ 70 \\ 0 \\ 0 \\ 0 \\ 80 \\ 0 \\ 0 \\ 0 \\ 0 \\ 90 \\ 80 \\ 70 \\ 70 \\ 0 \\ 70 \\ 70 \\ 0 \\ 0 \\ 0 \\ 70 \\ 0 \\ 0 \\ 0 \\ 0 \end{bmatrix}.$$

The first column of $X$ represents time remaining in terms of DCPs, $d_j$, and the second column represents the generated proxy for the remaining inventory, $x$.

---

**Algorithm 1** Observation Building for Data-Driven Bid Price Generation

---

**input:** historical pricing and time-to-departure data, $\boldsymbol{P}$ and $\boldsymbol{T}$,
capacity bound, $C$, set of flights $\boldsymbol{I}$, set of DCPs, $D = [d_1, d_2, \dots, d_{|D|}]$

set $\tilde{P} \leftarrow [\,]$
**for** $i \in \boldsymbol{I}$ **do:**
  set $\tilde{P}^i \leftarrow [\,]$
  **for** $d_j \in D$ **do:**
    set $P^i_{\leq d_j} \leftarrow \{P^i_n \mid T^i_n \leq d_j\}$, $\tilde{P}^i_{\leq d_j} \leftarrow [\,]$
    $\boldsymbol{E} \leftarrow \emptyset$
    **for** $k = 1: |P^i_{\leq d_j}|$ **do:**
      $\tilde{P}^i_{\leq d_j}(k) \leftarrow \max\left(P^i_{\leq d_j} \backslash E\right)$
      $E \leftarrow E \cup \left\{\max\left(P^i_{\leq d_j} \backslash E\right)\right\}$
    **end for**

**for** $k = |P^i_{\leq d_j}| + 1 : C$ **do:**

$\tilde{P}^i_{\leq d_j}(k) \leftarrow 0$

**end for**

$\tilde{P}^i \leftarrow [\tilde{P}^i;\ \tilde{P}^i_{\leq d_j}]$

**end for**

**end for**

In Algorithm 1, $|v|$ is used to denote cardinality and $v(i)$ is used to denote the $i^{th}$ element of a vector $v$. Additionally, $[A; B]$ represents row-wise concatenation of matrices $A$ and $B$.

Once the transformed observations are built from historical booking data, the next step is feeding that input data into the estimation algorithm, which we describe next.

#### *3.3.2.2 Deep Neural Network Algorithm for Bid Price Prediction*

We use a deep neural network for estimation. Given transformed booking data includes a proxy of bid price as a function of remaining capacity and time-to-departure (DCP) across several flights, the objective of predictive model is to come up with an unbiased estimation of bid prices for any given remaining inventory level and time-to-departure. Hence, the objective of the neural network model is to learn $\hat{b}(\hat{x}, d_j)$, given features $\hat{x}$ and $d_j$. We will refer to the bid price generated from our data-driven methodology as $b^{DD}(x, t)$ to differentiate it from the optimally generated bid price $b^*(x, t)$ introduced in Section 3.2.

It is worth noting that our methodology is in principle not limited to the use of neural networks to estimate bid prices. Indeed, the essence of the methodology lies in the transformation of historical prices into bid price proxies, which is the observation building step discussed in Section 3.3.2.1 Observation Building. It is possible to replace the deep neural network model with a simpler regression model especially for the simplified setting we discuss in the scope of this paper. However, as we consider more realistic extensions as discussed in Section 3.3.2.3 Practical Extensions, having a more sophisticated algorithm is preferrable. In addition to the increased predictive power, the neural network model provides many other advantages, which include flexibly handling categorical data, feature interactions, and additional covariates.

Note that our methodology is basically an ex-post greedy optimization process on historical data, which comes with certain advantages. These include not requiring an estimate of demand or demand distribution, not requiring any assumptions on demand, not limiting prices to come from fare classes or buckets, and finally imposing lighter requirements on data with no dependence on the historical availability or control data. Our approach is also limited to actual transactions and does not attempt to extrapolate and estimate demand for hypothetical scenarios, such as a lower or higher price than what is actually being offered.

#### *3.3.2.3 Practical Extensions*

In this section, we briefly discuss some of the extensions to the simplified setting above, and how they would impact our proposed methodology:

i. *Nonstationary demand:* In a practical setting, it is likely that the demand is non-stationary with respect to departure dates. That can be handled by modeling seasonality of departure day or week using Fourier series representations, and introducing these to data as additional covariates. Similarly, one can include additional features for day-of-week, time-of-day, etc. as applicable. Those additional features would simply be incorporated as additional covariates into the transformed input data, $X$. Trending can also be modeled similarly using linear, steps, or any other relevant basis functions, again in terms of departure day or week.

ii. *Batch arrivals:* It is possible to extend the model to handle group sales as discrete batch arrivals by repeating price data for the range of quantity covered, e.g. two units booked at price $p$ is equivalent to two occurrences of a unit sold at price $p$. For more continuous settings, such as booking by weight or volume (e.g., in cargo RM), the discrete capacity assumption can be relaxed by dividing total capacity into a number of buckets of different sizes with a common smallest unit of quantity. To give an example, a capacity of 1000 kg can be divided into buckets of [50, 100, 200, 500, 750, 1000] with a unit weight of 50 kg. Similar to the description in Section 3.3.2.1 Observation Building, the historical prices would be ordered in decreasing order and padded with zero for any remaining surplus of capacity. Subsequently, an additional step is needed to map these prices to the pre-defined buckets by a simple weighted averaging. These buckets would replace the column for remaining capacity proxy in $X$.

iii. *Network setting:* Even though we primarily cover the single leg (or single resource) setting in this paper, our method is relevant to the network RM setting as well. In that case, as the historical prices are at the product level (e.g., origin/destination/itinerary) and the bid price to be generated is at the resource level (e.g., flight leg), a preprocessing step is needed to prorate the historical prices to resource level. Common fare proration techniques from literature and practice still apply (McGill & Van Ryzin, 1999; Lapp & Weatherford, 2014), followed by the observation building step. The existence of multiple resources naturally manifests itself as additional covariates in $X$.

# 4. Numerical Study and Results

In this section, we present results from our numerical study where we test the performance of our method compared to an optimal benchmark which assumes perfect knowledge of the underlying true demand model and parameters associated with it. We use simulations that mimic the single-leg airline setting under the context of dynamic pricing with an exponential demand model as given by Equation 4 in Appendix A: Optimal Price under Exponential Demand Assumption The assumption of exponentially distributed willingness-to-pay is not a prerequisite for our method, but was chosen because it is commonly used in the RM literature, e.g. (Talluri & Van Ryzin, 2004; Gallego & Van Ryzin, 1994; Kumar, et al., 2021). We model a stationary arrival process with a constant expected arrival rate, $\lambda$, and expected WTP, $\alpha$. In order to evaluate our algorithm's performance with respect to different demand intensity scenarios we randomly set the arrival rate, $\lambda$, uniformly across a pre-defined range. For each scenario with a fixed $\lambda$, we run an initial baseline simulation implementing the optimal pricing policy which is a function of the optimal bid price (see Equation 6) to generate the historical booking data to be utilized as input to Algorithm 1 described in Section 3.3.2.1 Observation Building. Next, we apply Algorithm 1, feeding in pricing and time-to-departure sequences, $\boldsymbol{P}$ and $\boldsymbol{T}$ extracted from the simulated baseline data. Using the transformed input data, $X$ and $Y$, we train the neural network model to create bid price estimate as a function of remaining capacity and DCP. We then run two sets of simulations for each $\lambda$ setting from there on as follows:

- Simulation with optimal bid price as calculated from DP methodology. Note that this benchmark relies on the assumption of perfect knowledge of demand and WTP parameters, which is an unattainable optimum in practice.

- Simulation with bid prices estimated from our proposed data-driven method, in lieu of the optimal bid price. The pricing equation (6) is still used, replacing optimal bid price, $b^*(x,t)$, with $b^{DD}(x,t)$, thus still assuming perfect knowledge of the mean WTP, $\alpha$.

Both simulations are fed the same arrival stream with identical arrival times and WTP sequences. Figure 1 illustrates the data flow and setting for baseline simulations.

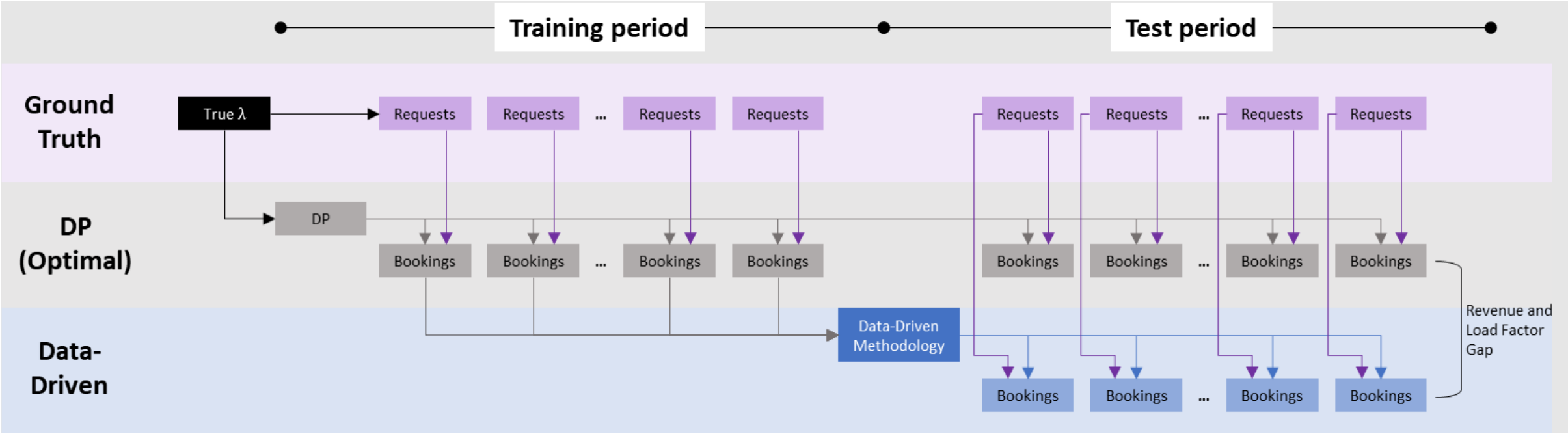

*Figure 1 Baseline Simulation Setting*

The baseline simulations help assess our proposed method's performance when trained with data generated by implementing the optimal policy, even though it has no knowledge of underlying parameters. In practice, the historical data used for training would almost always have been generated under a sub-optimal policy, and/or under a different underlying demand scenario. To test and compare our method's adaptivity and robustness, we next generate a set of simulations where the training and testing occurs under different arrival rate ($\lambda$) settings. Each of these trials is constructed by randomly combining two demand scenarios from the baseline simulations. One is used for training with a corresponding $\lambda_{train}$; and the other one used for test with a corresponding $\lambda_{test}$. From there on, three sets of simulations are run for each $(\lambda_{train}, \lambda_{test})$ tuple:

- Simulation with data-driven bid price: Simulation with arrival rate of $\lambda_{test}$, implementing bid prices generated from the data-driven methodology trained under the $\lambda_{train}$ setting.
- Simulation with bid price from a mis-specified DP (benchmark simulation): Simulation with arrival rate of $\lambda_{test}$, implementing optimal bid prices calculated for the $\lambda_{train}$ setting using DP. This corresponds to using a dynamic program with an inaccurate demand forecast.
- Simulation with optimal bid price: Simulation implementing optimal bid price which corresponds to $\lambda_{test}$ setting. Note that this is again equivalent to assuming perfect knowledge of demand, which is unrealistic in practice. We use this as an upper bound to help interpret the performance of the other two approaches.

We ensure all the simulations for each scenario with a unique $(\lambda_{train}, \lambda_{test})$ tuple is subject to the same arrival stream. Our objective with this second set of simulations is to test how well the bid prices we generate using our method perform when the true demand differs from the assumption (or in this case, the underlying assumption of historical data) used for training, which is often the case in practice. The scenarios with a significant change in underlying demand such that $\frac{\lambda_{test}}{\lambda_{train}} \gg 1$ or $\frac{\lambda_{test}}{\lambda_{train}} \ll 1$ , would be considered as demand shocks. However, our experiments cover more than only the extreme demand shocks to evaluate our methodology across a wider range of demand misspecification. We use optimal simulation results (representing the highest attainable expected revenue) to evaluate how well our methodology adjusts to the change in demand, in the absence of re-training. We benchmark our methodology's performance against the simulations with the bid price calculated by the DP for the original expected arrival rate, $\lambda_{train}$. We provide a flowchart for the simulation setting in Figure 2.

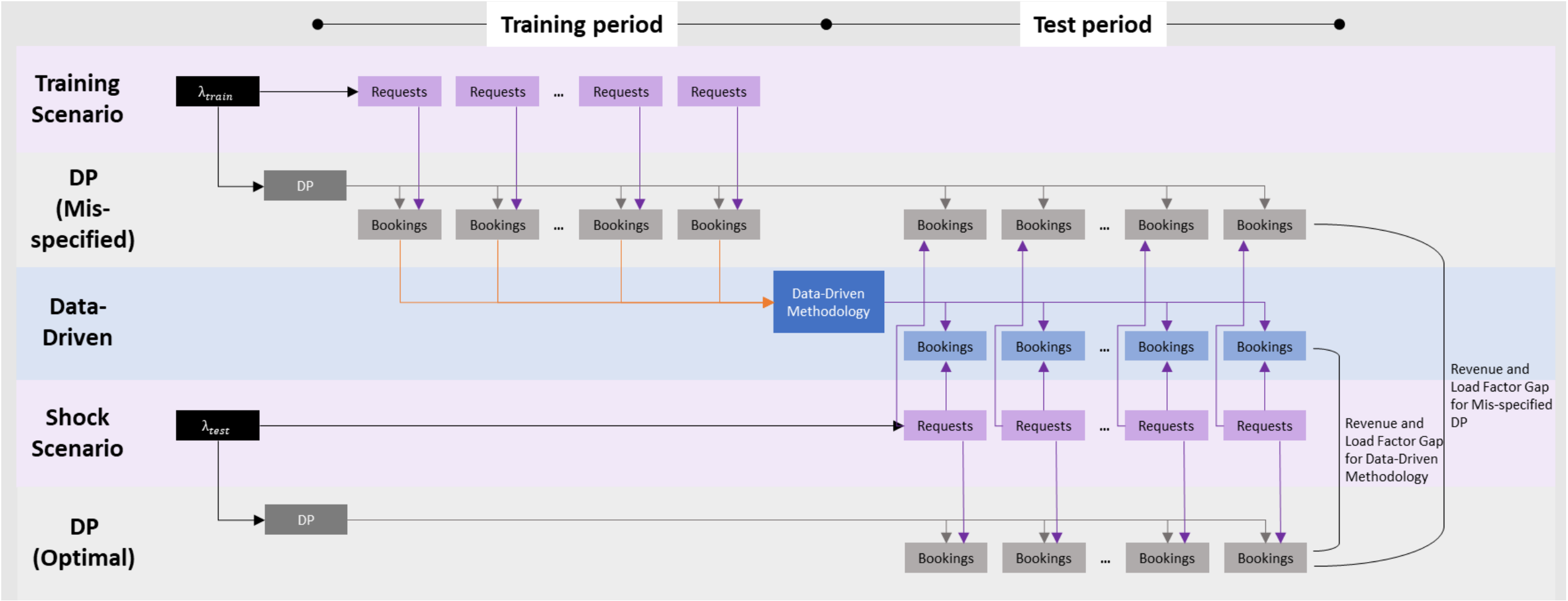

*Figure 2 Robustness Simulation Setting*

We will present more detailed descriptions of both numerical studies and results next.

## 4.1 Baseline Simulation and Results

### 4.1.1 Simulation Setting

For baseline simulations, we model a single flight leg across 300 departure dates (300 flights, $\boldsymbol{I} = \{1, \ldots, 300\}$) with a constant seat capacity set at 100. Note that departure date here is just a representation of an occurrence, as demand and WTP is assumed to be $i.i.d$ across departure dates. Both arrival rate and WTP are assumed to be stationary across the booking horizon, which is set as 300 days. For each scenario we randomly select the arrival rate per day, $\lambda$, uniformly from the range $[2.4, 3.6]$. Note that this is a very high utilization setting with an average load factor greater than 85%, where load factor is defined as the total booked seats divided by total capacity. This will ensure the bid prices are high enough to significantly matter in pricing. We group days priors into 10 uniform groupings such that DCPs used in Algorithm 1 are given by $D = [299, 269, 239, \ldots, 29]$. Finally, estimated bid prices, $b^{DD}(x, d_j)$, $x \in \{1,2, \ldots, 100\}, d_j \in D$ are interpolated between DCPs to get a separate bid price vector for each days prior in $\{0,1, \ldots, 299\}$, $b^{DD}(x, t)$. This is consistent with the granularity of the optimally calculated bid price, $b^*(x, t)$.

### 4.1.2 Neural Network Model Used in Estimation

We utilize a deep neural network model to estimate bid prices post observation building, using python's *hyperopt*[2] library (Bergstra, Yamins, & Cox, 2013) to optimize the neural network structure and hyperparameters. For tuning, we use observations generated from a single simulation scenario's data as described in Section 4.1.1 Simulation Setting. We then split the data by departure dates using an 80:20 split for training vs validation. **Table 1** lists the parameter space used in tuning as well as the parameters picked by the process. We utilize early stopping with a patience of 5 epochs, and use 'adam' as the optimizer.

[2] https://pypi.org/project/hyperopt/

***Table 1** Parameter space used for tuning of the deep neural network model, and selected parameters from the tuning process*

| Parameter/setting | Space | Selected Value |
|---|---|---|
| Number of layers (excluding the output layer) | [2, 3, 4] | 3 |
| Units per layer (set independently for each layer) | [4,8,16,32,64,128,256, 512] | [512, 8, 32] |
| Batch size | [32,64,128,256,500] | 128 |
| Learning rate | [0.001, 0.01, 0.1] | 0.001 |
| Regularization rate | [0,0.001, 0.01, 0.1] | 0.001 |
| Activation for the output layer | ['relu', 'softplus'] | 'softplus' |

### 4.1.3. Baseline Simulation Results

We run 100 simulation scenarios with a randomly set arrival rate of $\lambda$ as described in Section 4.1.1 Simulation Setting. The considered range of arrival rate ensures a high load factor, as shown in Figure 3. It illustrates the average optimal load factor over 300 departure dates corresponding to the arrival rate $\lambda$ for 100 simulations. Note that even though $\lambda$ is generated uniformly across the range of [2.4,3.6], the resulting load factors cluster around relatively higher values, as there is a cap to the utilization that can be reached as we keep increasing $\lambda$.

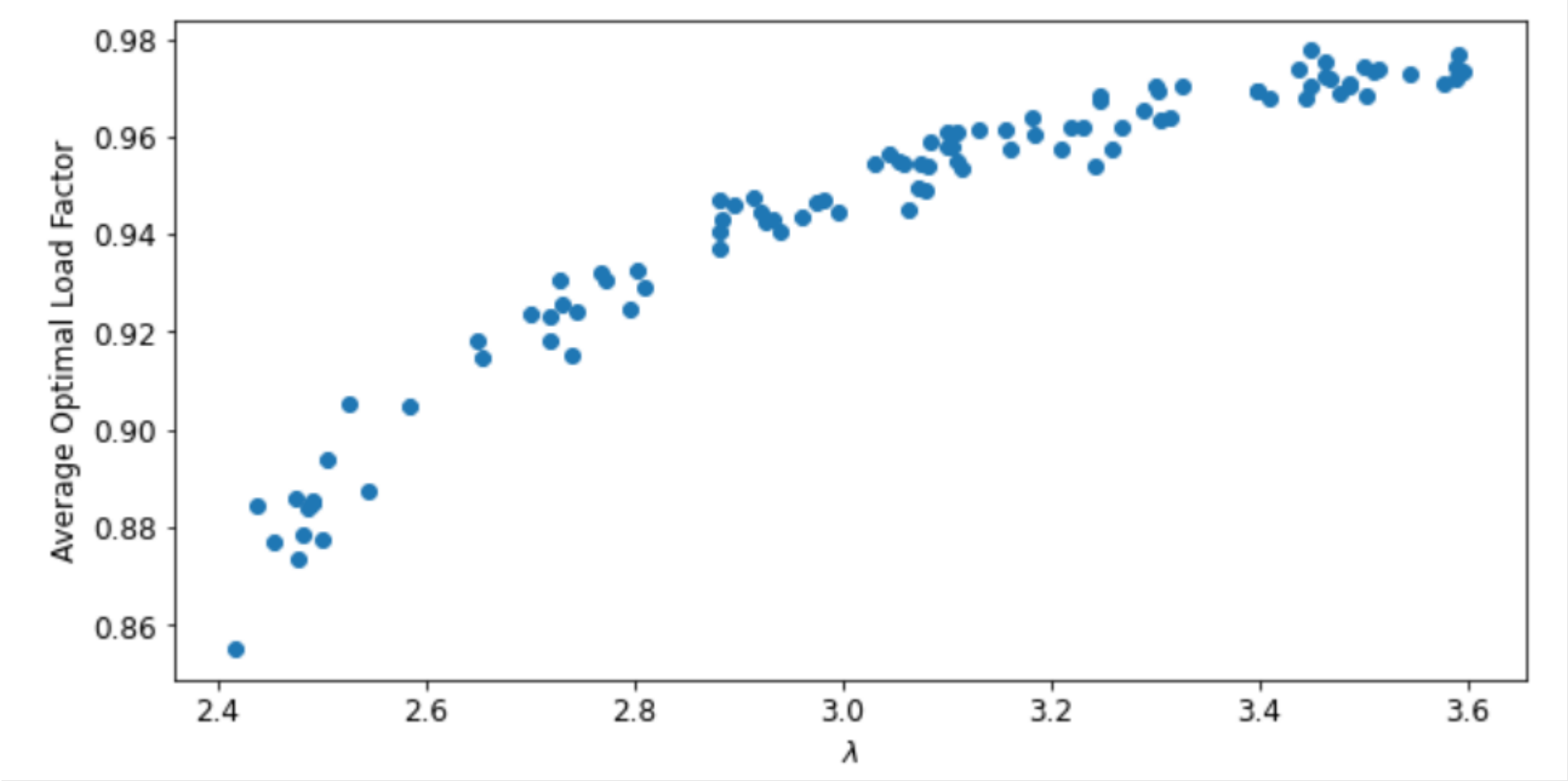

*Figure 3 $\lambda$ vs. Average Optimal Load Factor for 100 Simulation Scenarios*

Figure 4 illustrates observations generated using Algorithm 1 for all 300 flights (shown as dots, colored by flight), along with estimations generated by the neural network model (shown as blue line), for a simulation scenario with $\lambda = 3.4$. Charts provided represent different time points in the booking horizon. Note that the zeroes embedded in the observations shift as we get closer to departure in line with the logic in Algorithm 1. This enables our algorithm to capture the opportunity cost around what we will refer to as operating point. Even though bid prices are estimated for all combinations of days-to-departure and remaining capacity, some combinations are unlikely to be used in practice. For example, right at the beginning of the booking horizon at 300 days-to-departure it is almost sure that the remaining inventory is near the initial capacity of 100. Hence the very right-tail of the blue curve shown in Figure 4(a) is the section that would be utilized most frequently. For the mid-cycle example in Figure 4(b), which shows bid prices at 150 days before departure, that operating point shifts to the remaining capacity range of somewhere between 45 and 55. Towards the end of the booking horizon the flight is typically almost sold out and part of the curve with very low remaining capacity becomes the most relevant. Note that the bid price curves generated are the steepest around these operating points, which causes bid prices to be highly reactive to any deviations in the bookings from what is expected. This property is one reason for the "robustness" of our methodology that we will discuss in Section 4.2 Robustness Simulations and Results. As a reference, the optimal bid prices are also plotted (in gray). Note the intersection/overlay of optimal and data-driven bid price curves around those operating points, even though the curves deviate from each other significantly for other capacity ranges. This is another important insight from Figure 4 that explains how well the bid prices generated with our data-driven approach perform in simulations.

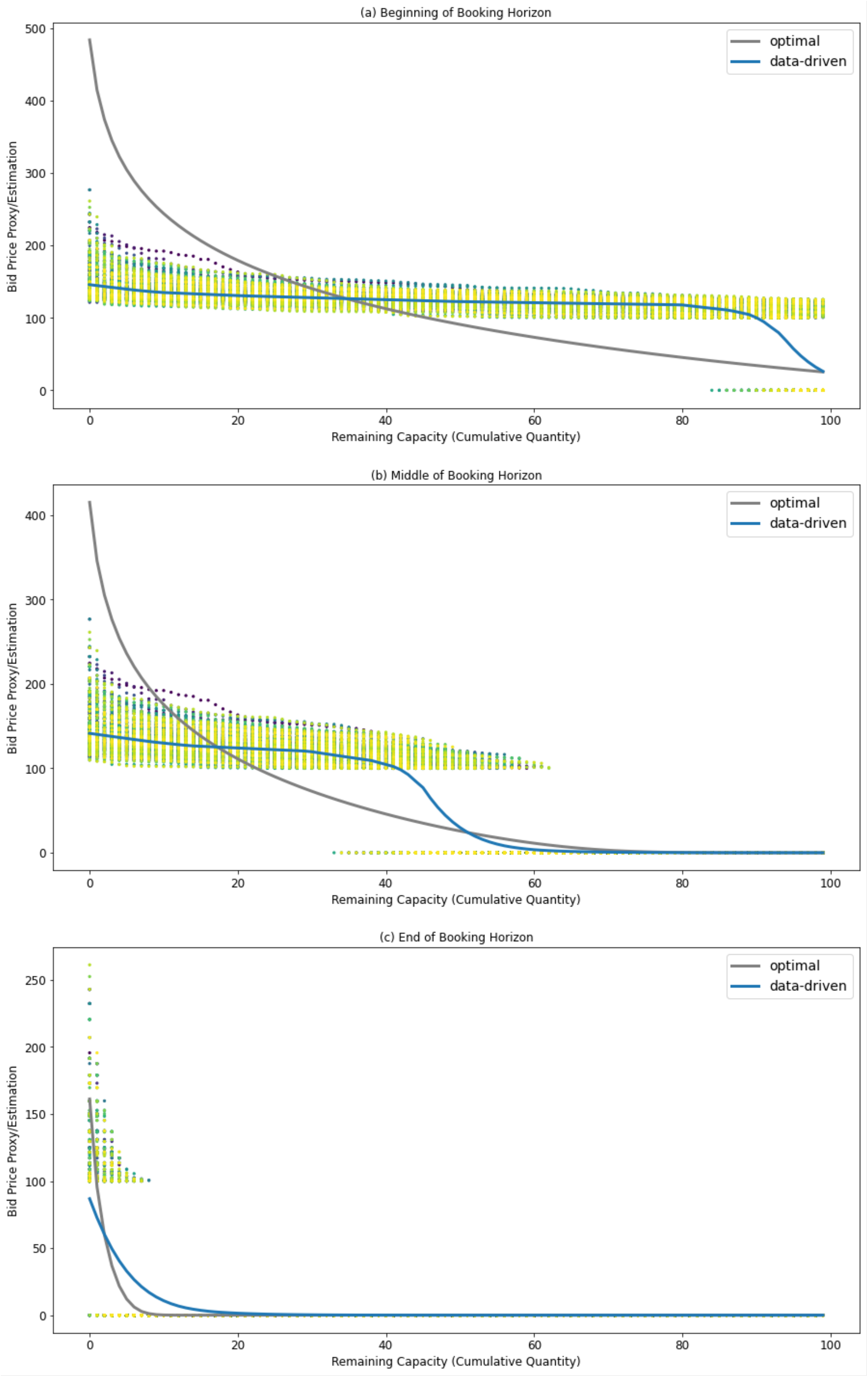

***Figure 4*** Bid Price Proxy from Observations of 300-day Sample ($\lambda = 3.4$) and Estimated (Data-Driven) and Optimal Bid Prices for (a) Days Before Departure, $dbd \leq 300$, (b) $dbd < 150$, (c) $dbd < 10$.

Figure 5 summarizes results across all simulation scenarios with varying $\lambda$. Each dot represents a demand scenario corresponding to a single arrival rate of $\lambda$. The horizontal axis shows the average load factor obtained over the 300 samples under the optimal bid price policy. The vertical axis shows the average relative revenue gap of the data-driven policy compared to the optimal one. Note that the relative revenue gap compared to the optimal worsens (decreases), as the load factor increases. This is consistent with the fact that bid prices correspond to a more significant portion of the total revenue as load factor increases and capacity gets more constrained and valuable. The relative gap from optimal stays above $-1.25\%$ across all simulation scenarios with an overall average of $-0.46\%$.

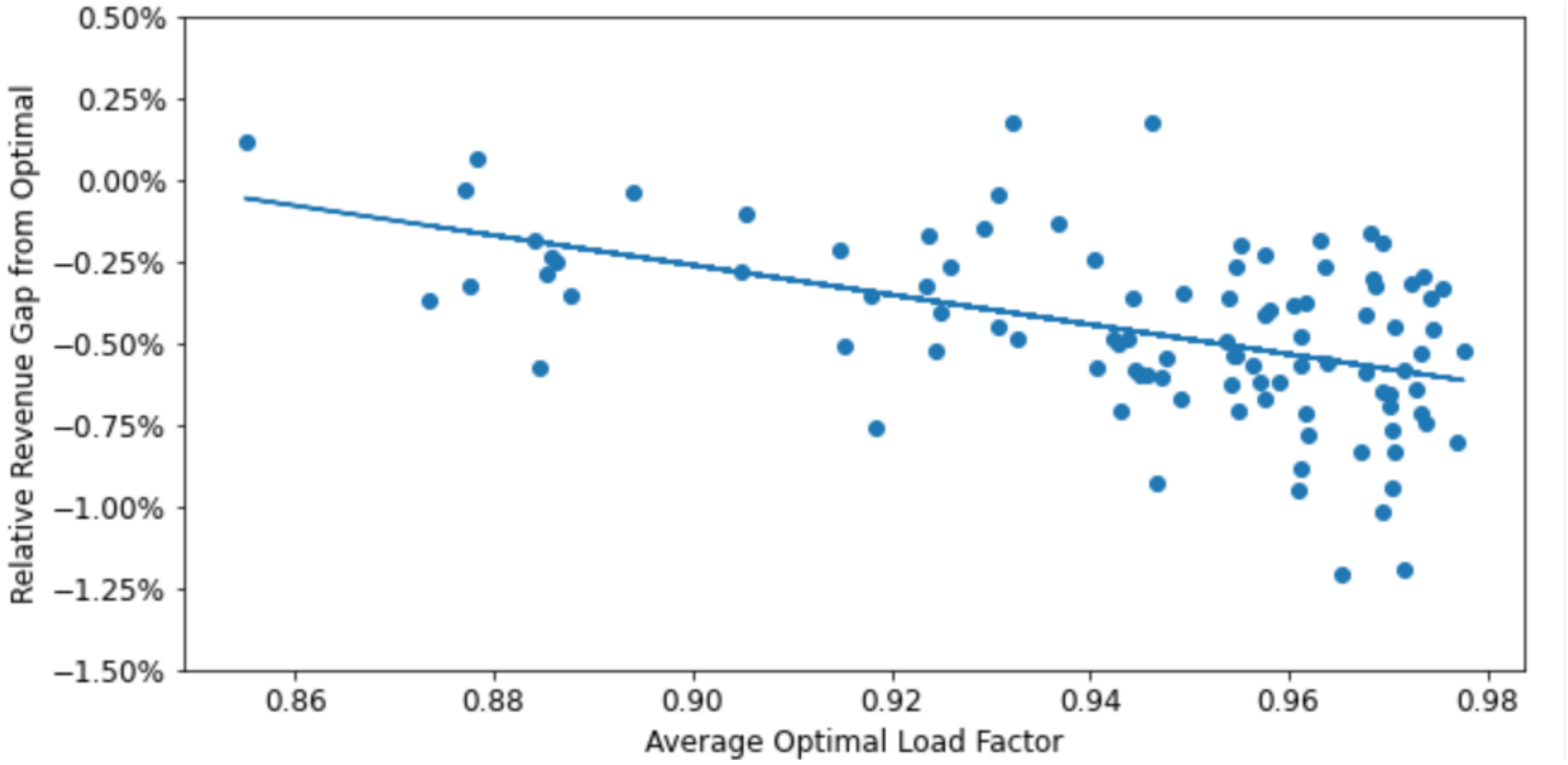

*Figure 5 Average Relative Revenue Gap from Optimal with respect to the Average Optimal Load Factor with Linear Trendline (Each Dot Represents a Simulation Scenario with a Given Expected Arrival Rate of $\lambda$ and 300 Flights)*

Figure 6 shows similar results in terms of load factor. Here, the values on the vertical axis represent the average difference between the load factor achieved with the data-driven method and the optimal DP policy. Our results show larger gaps in load factor compared to revenue with respect to optimal. However, the average gaps in load factor still falls above $-5\%$ overall.

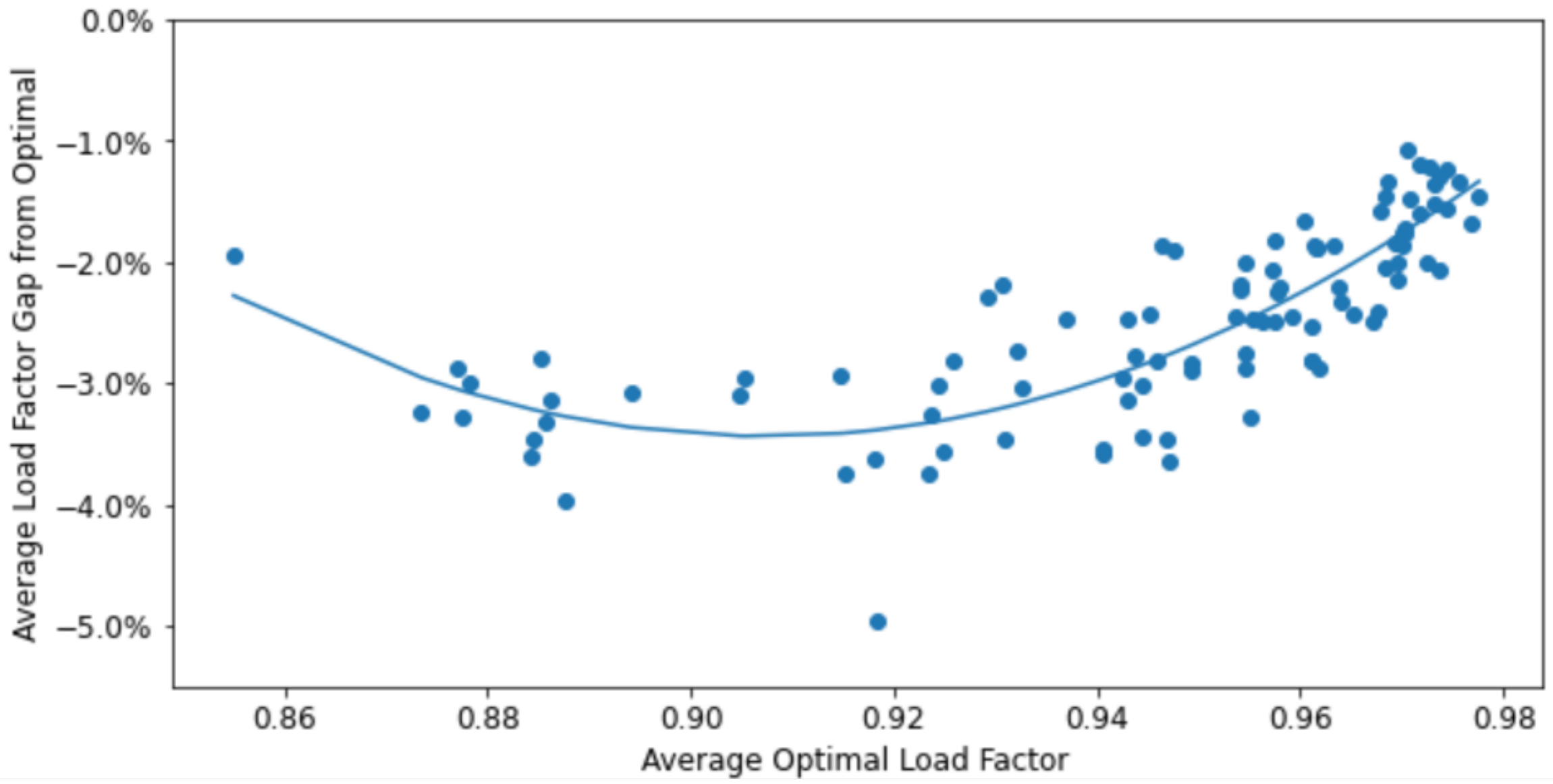

*Figure 6 Average Load Factor Gap from Optimal with respect to the Average Optimal Load Factor with Polynomial Trendline (Each Dot Represents a Simulation Scenario with a Given Expected Arrival Rate of λ and 300 Flights)*

## 4.2 Robustness Simulations and Results

### 4.2.1. Simulation Setting

For this second set of simulations, we randomly sample from the baseline simulations to pick a $\lambda_{train}$ and $\lambda_{test}$, as described in the introduction of Section 4. To explore a wider range of scenarios, we uniformly sample $\lambda_{train}$ from the range of $[2.4,3.6]$ and $\lambda_{test}$ from $[1.8,3.6]$, which means the ratio of arrival rates per test and training period, $\frac{\lambda_{test}}{\lambda_{train}}$ fall into the range $[0.5,1.5]$. We train our neural network model with the simulated data corresponding to $\lambda_{train}$ setting and feed the estimated bid prices into a new simulation with an expected arrival rate of $\lambda_{test}$. We compare our method's performance to that of the simulations with bid prices generated from a DP assuming an expected arrival rate of $\lambda_{train}$. We compare both simulations to the optimal solution with perfect knowledge of $\lambda_{test}$, which is practically an unattainable optimum. The objective is to evaluate how well the bid prices from our methodology and the DP-based approach adjust when applied in an unexpected demand setting without any re-calculations or re-training. We run a total of 500 scenarios each representing a single $(\lambda_{train}, \lambda_{test})$ tuple. For each scenario, we simulate 500 flights each with a randomly generated arrival stream. The same arrival stream is used for data-driven, benchmark (mis-specified DP), and optimal (optimal DP) simulations.

### 4.2.2. Simulation Results

Figure 7 summarizes the distribution of the test and training arrival rate (demand) ratio, $\frac{\lambda_{test}}{\lambda_{train}}$, across the 500 simulations. A ratio of 1 means that demand is exactly as expected, while values far below or above 1 represent much lower or higher demand than expected respectively.

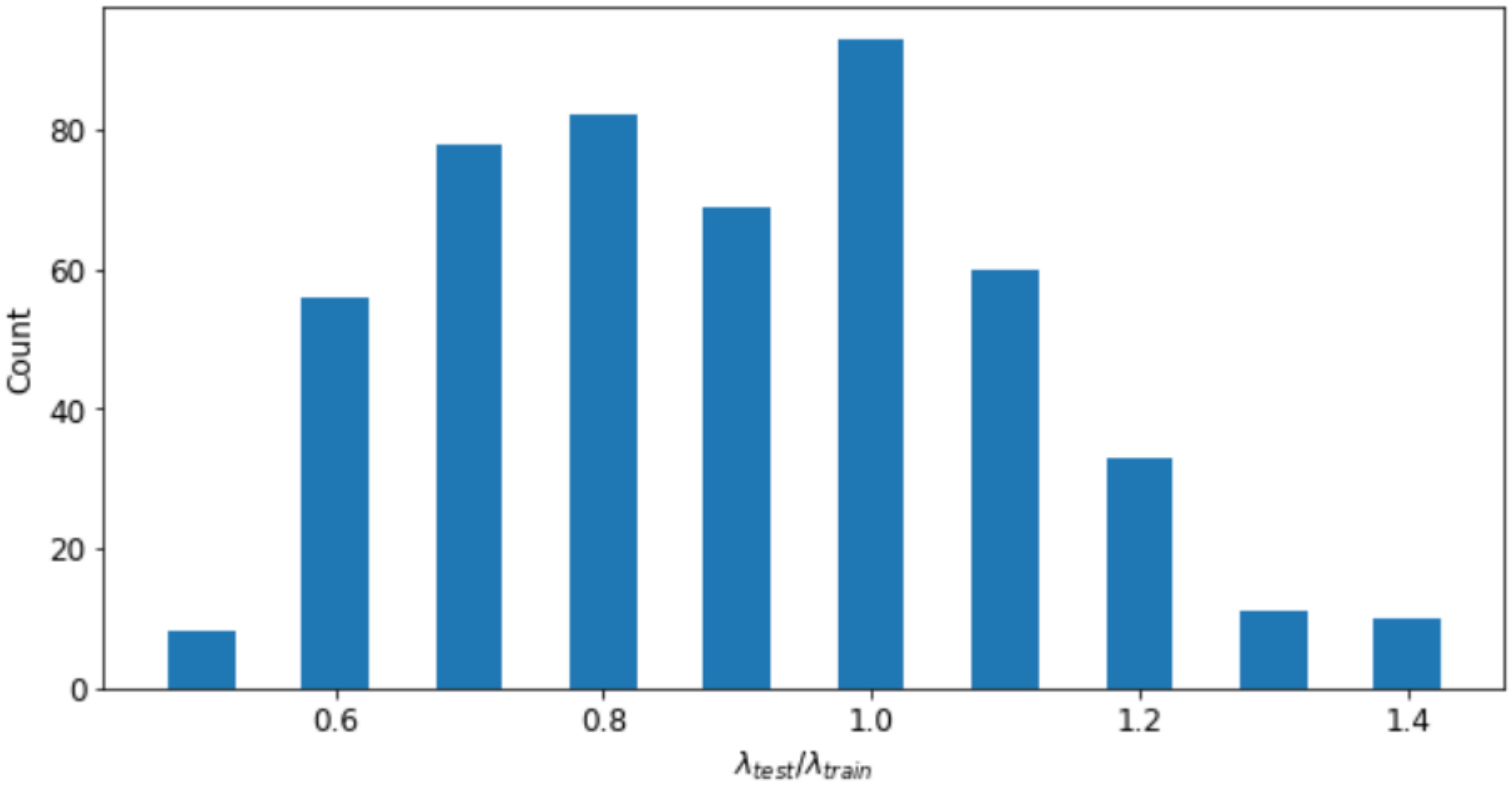

*Figure 7 Number of Simulations per $\lambda_{test}/\lambda_{train}$ (Rounded to First Decimal).*

Comparing each simulation's revenue and load factor results to those of the optimal counterpart, we run similar analysis as in Section 4.1 Baseline Simulation and Results to evaluate the performance of both the data-driven and the benchmark DP simulations. Figure 8 illustrates the revenue gap results. As expected, DP simulations perform very close to their optimal counterparts, when $\lambda_{test}/\lambda_{train} \approx 1$. This is simply because when $\lambda_{test} = \lambda_{train}$, DP methodology would generate the optimal bid prices, which means benchmark and optimal simulations are equivalent. However, when demand is significantly mis-specified, especially when $\lambda_{test} \gg \lambda_{train}$, the revenue gap quickly grows very large. On the other hand, data-driven simulations seem to always perform slightly worse than the optimal DP solution, but the gap seems to be capped ($< 1\%$ for the studied range) even for large demand shocks. When $\lambda_{test} \ll \lambda_{train}$, both methods behave similarly to each other and are both close to optimal. This is because for very low demand scenarios the optimal bid price is usually zero, which is easy to find for all methods. Figure 9 shows similar results in terms of load factor gaps from optimal. For both the data-driven method and the DP simulations, load factor seems to be impacted more strongly by demand misspecification. Similar to the patterns observed for the revenue, the benchmark DP solution is close to optimal when $\lambda_{test}$ is close to $\lambda_{train}$, while load factors generated by data-driven bid prices stay relatively closer to the optimum when $\lambda_{train} \gg \lambda_{test}$ or $\lambda_{train} \ll \lambda_{test}$.

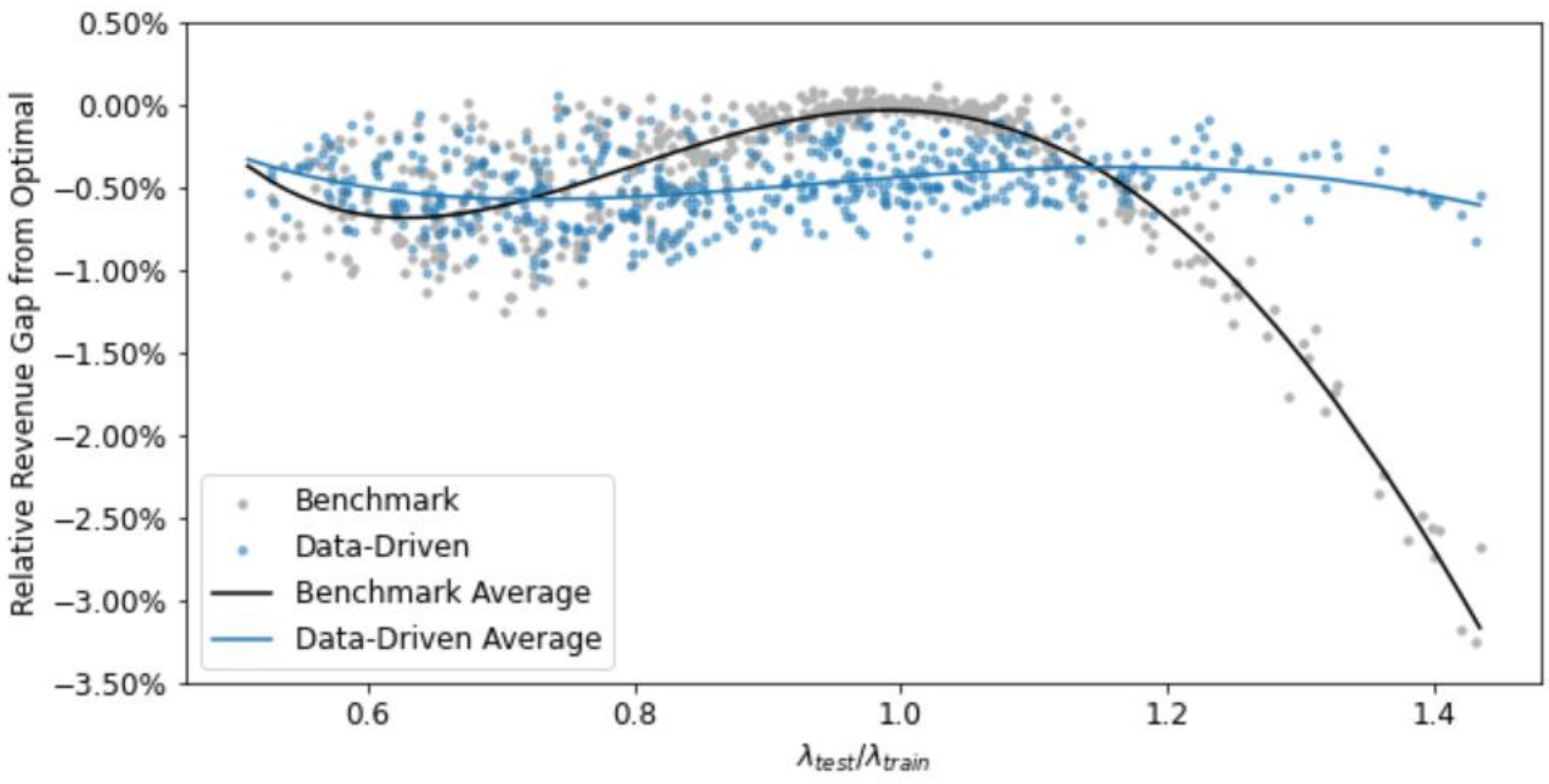

*Figure 8 Average Revenue Gap from Optimal with respect to the Ratio of $\lambda_{test}$ to $\lambda_{train}$ for Benchmark and Data-Driven Simulations with Their Respective Trendlines (Each Dot Represents a Simulation Scenario with a Given $\lambda_{test}/\lambda_{train}$ and 500 Flights).*

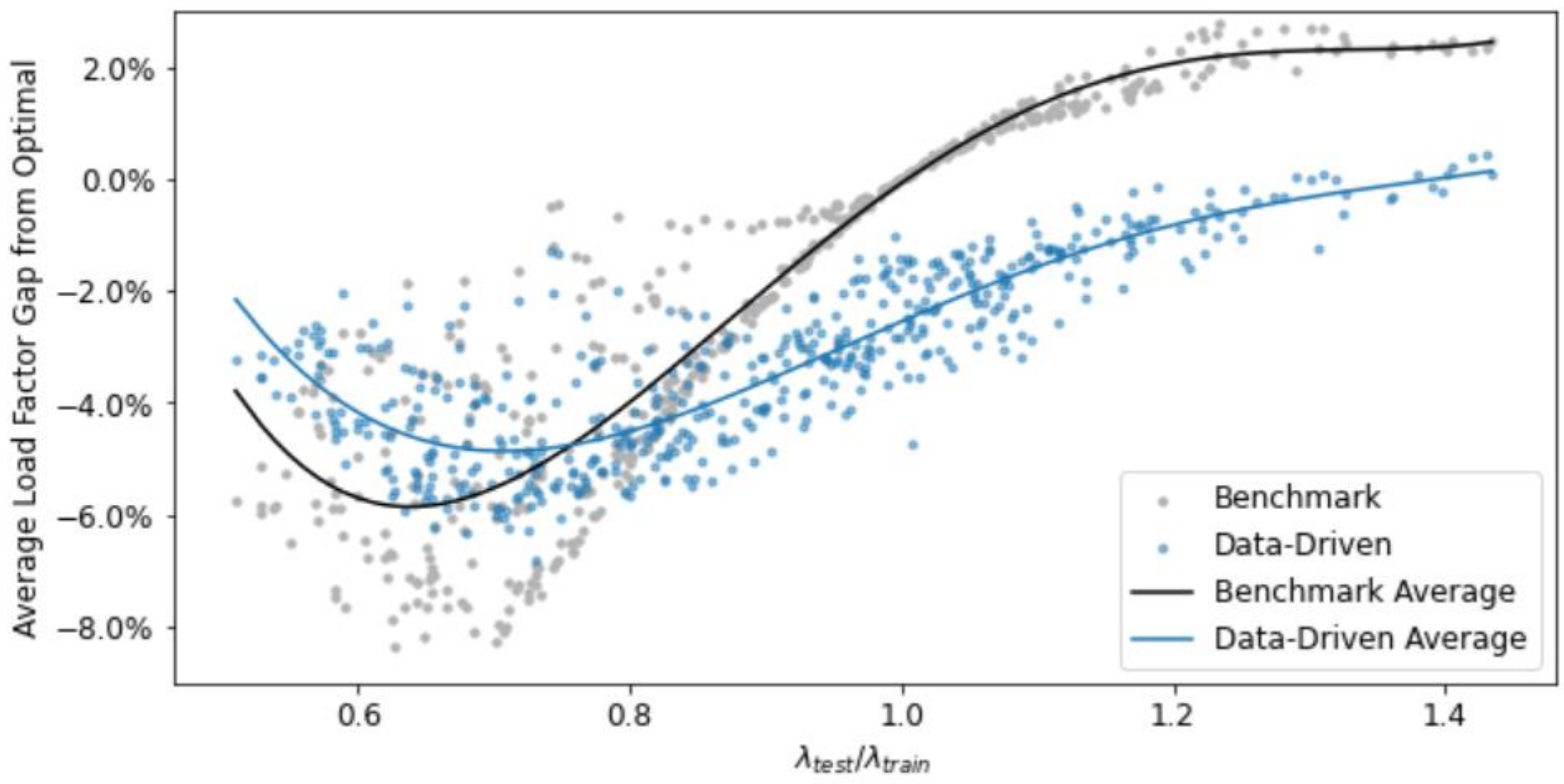

*Figure 9 Average Load Factor Gap from Optimal with respect to the Ratio of $\lambda_{test}$ to $\lambda_{train}$ for Benchmark and Data-Driven Simulations with Their Respective Trendlines (Each Dot Represents a Simulation Scenario with a Given $\lambda_{test}/\lambda_{train}$ and 500 Flights).*

Figure 10 summarizes results in terms of the ratio of data-driven to benchmark simulations with respect to the different demand ratios, $\lambda_{test}/\lambda_{train}$. In agreement with Figure 8, the total revenues of data-driven and benchmark simulations are comparable to each other when $\lambda_{test} \leq \lambda_{train}$. Other than the observation that our data-driven method is doing slightly worse when $\lambda_{test} \approx \lambda_{train}$, it is doing noticeably and increasingly better as $\lambda_{test}$ becomes larger compared to $\lambda_{train}$. Load factor ratios show larger variance in general with our data-driven methodology exhibiting larger load factors as $\lambda_{test} \ll \lambda_{train}$ and lower load factors otherwise.

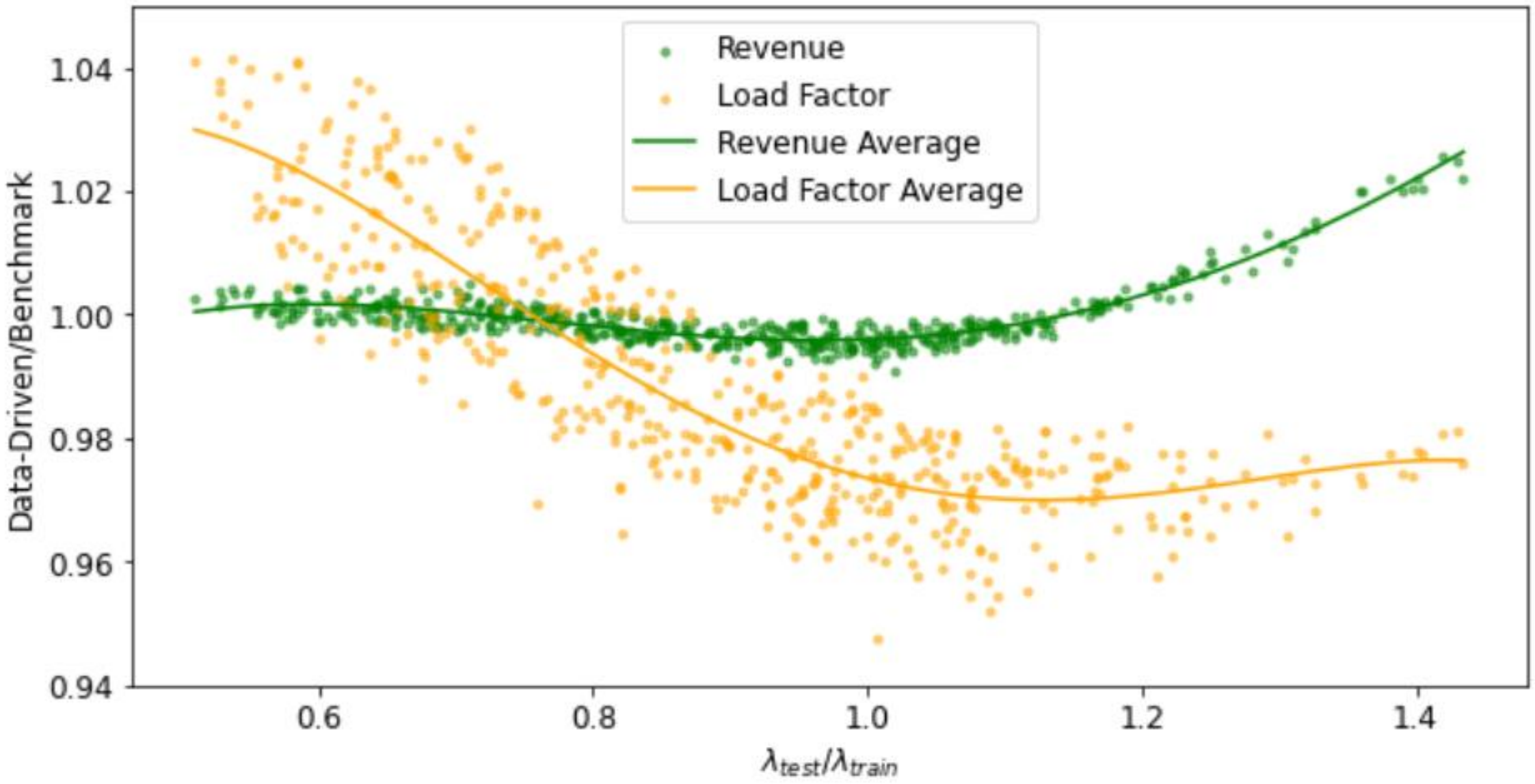

*Figure 10 Ratio of Total Revenue and Average Load Factor of Data-Driven to Benchmark Simulations with respect to the Ratio of $\lambda_{test}$ to $\lambda_{train}$ with Their Respective Trendlines (Each Dot Represents a Simulation Scenario with a Given $\lambda_{test}/\lambda_{train}$ and 500 Flights).*

In summary, our data-driven methodology is robust in revenue attainment. It is only slightly worse than the benchmark DP methodology when demand estimation is close to perfect. However, it does significantly better when demand is much higher than expected, preserving its low gap to optimal revenue. Underlying this robustness is indeed the shape of the bid price curves generated by our data-driven approach. As shown in Figure 4, the data-driven and optimal bid price vectors intersect around operating points with data-driven bid price vector having a much steeper slope. This implies that it reacts more strongly to any deviations from expectation. If bookings arrive more quickly than expected, the data-driven methodology reacts by increasing bid price more rapidly compared to the DP-calculated bid price curve, which would slow down bookings in response. Similarly, if the bookings arrive more slowly than expected, data-driven methodology reacts by decreasing the bid price more rapidly compared to the DP-calculated bid price which would result in acceptance of more bookings in response. This implicit feedback loop to aggressively push bookings towards the expected booking curve is the reason for the robustness of our data-driven methodology when applied under an unexpected demand shock.

## 5. Concluding Remarks and Future Work

In this paper, we presented a data-driven approach to generate bid prices, which are commonly used in RM. Unlike conventional RM techniques, our methodology skips demand forecasting, does not make any assumptions on demand, and does not incorporate any mathematical optimization techniques. It utilizes solely historical booking data and turn them into a proxy of bid prices via a greedy heuristic and uses a neural network model for estimation.

Our method's flexibility in terms of data requirements, inherent assumptions and ease of implementation makes it a great candidate for industries that have not established a matured RM practice yet. Moreover, our method is a great fit for industries that face severe demand volatility because it requires not demand forecasting and the bid price predictions are robust to mis-specified demand. It is very promising that the simulation results show only a small gap to optimum under significant deviations between assumed and true demand, which occur frequently in the practice of RM.

The immediate extensions to the studied problem are already listed in Section 3.3.2.3 Practical Extensions Another extension to our methodology would be incorporating an explicit feedback loop in addition to the inherent adaptivity of the steep bid price curve around the operating points. This could improve the performance of our method further. Another area of study is the inclusion of additional data sources as covariates (e.g., economic indicators, weather forecast, etc.) that could further improve bid price predictions. Incorporation of additional covariates is relatively cheap while using a machine learning algorithm. In addition to all the flexibility it brings, our method is also scalable given its simplicity. Therefore, it is promising both as a pragmatic approach widely usable in practice, as well as a starter of a new research avenue under the umbrella of recently trending data-driven RM.

# Acknowledgements

The authors thank Michael Wu for his editing and feedback that led to considerable improvements in the presentation of this paper.

# Appendix A: Optimal Price under Exponential Demand Assumption

Continuing from Section 3.2, as an illustrative example, under exponentially distributed WTP, the purchase probability at a price, $P_w(p,t)$, is given by

$$P_w(p,t) = e^{-\frac{(p-p_0(t))}{\alpha(t)}} \tag{4}$$

where, the mean willingness-to-pay is given by $\alpha(t)$. From the DP formulation in Equation (2), the optimal price equation is given by

$$p^*(x,t) = \underset{p}{\operatorname{argmax}}\, P_w(p,t)\,.\,(p - b(x,t-1)) \tag{5}$$

Therefore, under the exponential demand assumption given by Equation (4), this turns into

$$p^*(x,t) = \max[p_0(t), \alpha(t) + b(x,t-1)]. \tag{6}$$

In this case, generation of $b(x,t)$ and $p^*(x,t)$ both require the estimation of the demand model parameters $\lambda(t), \alpha(t),$ and $p_0(t)$.

# Appendix B: Numerical Example for Relation of Data-Driven Approach to EMSR

Let's look at an example comparing EMSR results to those of the data-driven approach with no consideration of booking times and using a simple average for estimation.

Let us denote the total demand for fare class 1 by $D_1$ and assume it is normally distributed, $D_1 \sim N(3,2)$, with a fare of $f_1 = 400$. The corresponding EMSRs for class 1 calculated by $f_1 P(D_1 > s)$ for each remaining capacity $s = \{1, \ldots, 10\}$ as described in Section 3.1 are provided in Table 2.

*Table 2 EMSRs calculated for* $D_1 \sim N(3,2)$ *and* $f_1 = 400$.

| seat | $EMSR_1$ |
|---|---|
| 1 | 336.54 |
| 2 | 276.58 |
| 3 | 200.00 |
| 4 | 123.42 |
| 5 | 63.46 |
| 6 | 26.72 |
| 7 | 9.10 |
| 8 | 2.48 |
| 9 | 0.54 |
| 10 | 0.09 |

Let's now look into our data-driven methodology under the same assumptions. Sampling from $D \sim N(3,2)$ to represent 5 flights (departure dates), applying algorithm 1 to pad with zeroes for any surplus capacity, and then using average for estimation of bid price $b^{DD}(x,t)$ yields results in Table 3. Results based on 100 samples are also provided in the same table.

*Table 3 EMSRs calculated for $D_1 \sim N(3,2)$ and $f_1 = 400$, with corresponding data-driven bid prices generated from sampled data.*

| fare | 400 | | |
|---|---|---|---|
| mean | 3 | | |
| Std Dev | 2 | | |
| seat | EMSR | $b^{DD}(x,t)$ with 5 samples | $b^{DD}(x,t)$ with 100 samples |
| 1 | 336.54 | 400 | 332 |
| 2 | 276.58 | 320 | 284 |
| 3 | 200.00 | 320 | 236 |
| 4 | 123.42 | 240 | 148 |
| 5 | 63.46 | 160 | 84 |
| 6 | 26.72 | 80 | 24 |
| 7 | 9.10 | 80 | 4 |
| 8 | 2.48 | 80 | 0 |
| 9 | 0.54 | 0 | 0 |
| 10 | 0.09 | 0 | 0 |

As seen by the results, $b^{DD}(x,t)$ approach EMSR values as the sample size increases.